\documentclass[nonacm]{acmart}
\usepackage{balance}
\AtBeginDocument{%
  \providecommand\BibTeX{{%
    \normalfont B\kern-0.5em{\scshape i\kern-0.25em b}\kern-0.8em\TeX}}}

\usepackage{amsmath}
\usepackage{xcolor}
\usepackage[framemethod=tikz]{mdframed}
\usetikzlibrary{shadows}

\definecolor{mygray}{HTML}{F5F5F5}

\begin{document}

\title{Automated Claim Matching with Large Language Models: Empowering Fact-Checkers in the Fight Against Misinformation }

\author{Eun Cheol Choi}
\email{euncheol@usc.edu}
\orcid{0003-0861-1343}

\author{Emilio Ferrara}
\email{emiliofe@usc.edu}
\affiliation{%
  \institution{University of Southern California}
  \streetaddress{3502 Watt Way}
  \city{Los Angeles}
  \state{California}
  \country{USA}
  \postcode{90089-0281}
}

\renewcommand{\shortauthors}{Choi and Ferrara}

\begin{abstract}
In today's digital era, the rapid spread of misinformation poses threats to public well-being and societal trust. As online misinformation proliferates, manual verification by fact checkers becomes increasingly challenging. We introduce FACT-GPT (\textit{Fact-checking Augmentation with Claim matching Task-oriented Generative Pre-trained Transformer}), a framework designed to automate the claim matching phase of fact-checking using Large Language Models (LLMs). This framework identifies new social media content that either supports or contradicts claims previously debunked by fact-checkers. Our approach employs \textit{GPT-4} to generate a labeled dataset consisting of simulated social media posts. This data set serves as a training ground for fine-tuning more specialized LLMs. We evaluated FACT-GPT on an extensive dataset of social media content related to public health. The results indicate that our fine-tuned LLMs rival the performance of larger pre-trained LLMs in claim matching tasks, aligning closely with human annotations. This study achieves three key milestones: it provides an automated framework for enhanced fact-checking; demonstrates the potential of LLMs to complement human expertise; offers public resources, including datasets and models, to further research and applications in the fact-checking domain.

\end{abstract}

\begin{teaserfigure}
  \includegraphics[width=\textwidth]{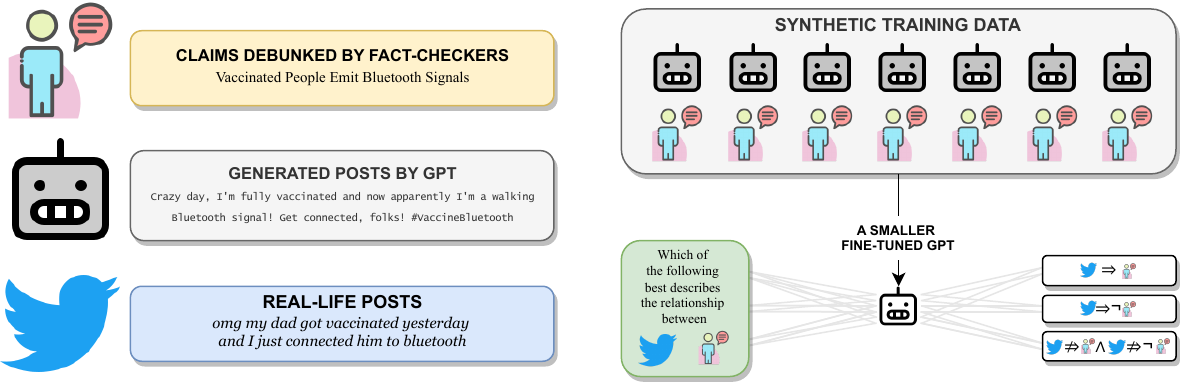}
  \caption{Overview of FACT-GPT, our framework aimed at augmenting fact-checkers' ability to debunk misinformation}
  \Description{An overview of claim matching system leveraging LLMs fine-tuned on labeled generated social media posts.}
  \label{fig:teaser}
\end{teaserfigure}


\maketitle

\section{Introduction}
Fact-checking serves as a vital tool in the fight against misinformation \cite{walter2020fact, nakov2021automated, augenstein2023factuality}. This process involves investigating the truthfulness of claims from public discourse and subsequently publishing the findings \cite{graves2019}. Given the rapid proliferation of misinformation on social media and other online platforms \cite{qiu2017, vosoughi2018}, there is an unprecedented need for timely and extensive fact-checking. However, the fact-checking process is complex and time-consuming, requiring multiple steps from identifying claims to making final conclusions \cite{hassan2017, elsayed2019}. Hence, it may be impractical or even unfeasible for fact checkers to manually verify every dubious claim that arises.

To augment human fact-checkers' capabilities, researchers are exploring the integration of artificial intelligence (AI) tools into the fact-checking pipeline. Yet, fully automating this process with AI poses risks, potentially undermining the journalistic norms and practices that underpin fact-checking \cite{nakov2021automated}. Therefore, the goal should not be to replace human expertise, but to augment human decision making. The concept of "augmented intelligence" \cite{degallier2022, ieee2023} provides a suitable framework for the development of AI that enhances the efficiency and consistency of fact checkers without compromising their principles.

This paper explores the potential use of large language models (LLMs) in helping the "claim matching" stage of the fact-checking process, a step where new instances of previously fact-checked claims are identified \cite{shaar2020}. This benefits practitioners by reducing redundant verification, online platforms by aiding content moderation, and researchers by analyzing misinformation from a large corpus. We evaluate various LLMs on their ability to judge the textual entailment between social media posts and verified claims. Our findings suggest that LLMs can reliably match claims, offering performance comparable to human ratings. If properly implemented, claim matching techniques could assist fact checkers in the early identification of recurring misinformation. This study is a first step in the direction of augmenting fact-checking work transparently with LLMs.

\section{Proposed Framework}

\subsection{Task Definition}

To evaluate the abilities of various LLMs, from proprietary to open-source models, in claim matching, we employ a \textit{textual entailment task} \cite{marelli2014}. Textual entailment classifies pairwise relationships into one of three categories: \textit{Entailment}, \textit{Neutral}, and \textit{Contradiction}. A pair is classified as 'Entailment' when the truth of Statement A implies the truth of Statement B. It is classified as 'Neutral' when the truth of Statement A neither confirms nor denies the truth of Statement B. Finally, a pair is marked as 'Contradiction' when the truth of Statement A implies that Statement B is false. Textual entailment tasks focus on everyday reasoning, not strict logic, so human judgement and common sense determine the ground truth \cite{marelli2014, pado2022}. Note that the sequence of statements in the task is crucial, as the entailment could be either unidirectional or bidirectional. In other words, the proposition ‘when A is true, B is also true’ is not equal to ‘when B is true, A is also true’.

We postulate that if a model excels at entailment tasks, it will also be reliable in claim matching. For example, if a pair consisting of a tweet and a false or misleading claim exhibits an entailment relationship, it can be inferred that the tweet is also spreading the same false or misleading claim. The entailment task is particularly applicable to declarative sentences, as it directly concerns the truth value of the pair \cite{bentivogli2016}. The entailment task has been previously used successfully in rumor detection as well \cite{yavary2019}.

\subsection{Collecting Debunked Claims}
Here, we focus on public health-related misinformation, in particular fact-checking misinformation about COVID-19, as case study.
False claims debunked by professional fact checkers were obtained from \textit{Google Fact Check Tools} (\url{https://toolbox.google.com/factcheck/explorer}) and \textit{PolitiFact} (\url{https://www.politifact.com/}). We collected claims from January 2020 through December 2021. We selected claims that had keywords like 'covid-19,' 'coronavirus,' or 'pandemic' from Google Fact Check Tools, and those categorized under COVID-19 from PolitiFact. Since this approach focuses on the textual content of the false claims, only claims meeting the following two criteria were included for analysis:

\begin{itemize}
  \item The claims did not refer to external images, videos, or URLs.
  \item Claims were unequivocally labeled false, incorrect, or fake.
\end{itemize}

\begin{figure}[t]
    \centering
    \includegraphics[width=0.45\textwidth]{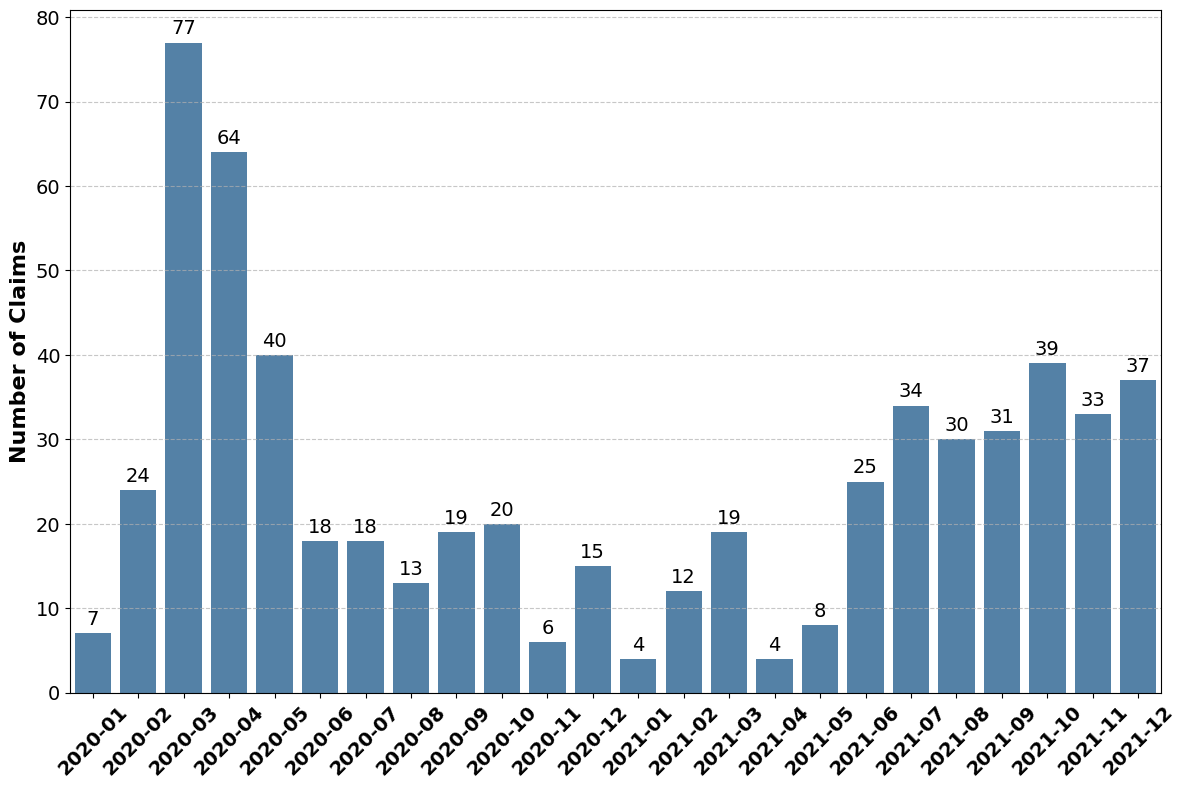}
    \caption{Monthly distribution of debunked claims}
    \label{fig:distrib}
\end{figure}

After removing duplicates, this process yielded a total of 1,225 false claims. Figure \ref{fig:distrib} shows the monthly distribution of claims used in this study.

\subsection{Constructing Test Data}
\begin{figure}[ht]
    \centering
    \includegraphics[width=0.45\textwidth]{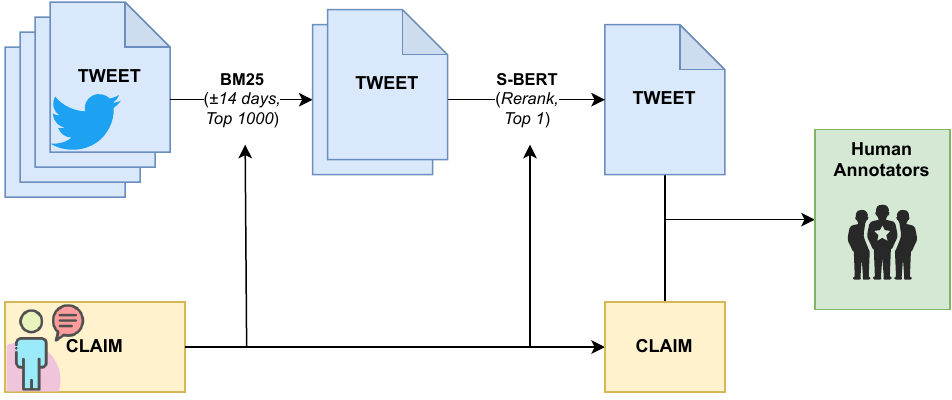}
    \caption{Workflow of test data construction}
    \label{fig:workflow}
\end{figure}

Figure \ref{fig:workflow} illustrates our workflow for the construction of test data. We first paired false claims debunked by fact-checkers with tweets from a Coronavirus Twitter dataset, considering token and semantic similarity. These pairs undergo human annotation via Amazon Mechanical Turk, where highly-rated workers classify them into categories of \textit{Entailment}, \textit{Contradiction}, or \textit{Neutral}. Majority vote is used to finalize these classifications, resulting in a curated test data set.

\subsubsection{Pairing Tweets with Claims}

In conducting the claim matching task, two types of data are paired: false claims verified by fact-checkers and a linguistic corpus from natural communication settings. For the latter, we used a public \textit{Coronavirus Twitter data set} collected from January 2020 to December 2021 \cite{chen2020}. These data consist of real-time tweets collected using the \textit{Twitter Streaming API}. Similarly to our approach with claim data, we selected only original tweets without URLs, images, or videos to focus on textual modality, resulting in 86,883,325 tweets.

To find tweets that match debunked claims, we employed two metrics: BM25 \cite{robertson1995okapi} for token similarity and Sentence-BERT (\textit{all-MiniLM-L6-v2}) for semantic similarity \cite{reimers2019}. This approach is consistent with previous literature \cite{hassan2017, shaar2020}, which considered both types of similarities for claim matching tasks. The top 1,000 best-matching tweets for each verified claim were initially retrieved based on BM25 scores within a $\pm$ 14-day window from the day the false claim was first made. These tweets were then reranked on the basis of the cosine similarity between the S-BERT embeddings of each verified claim and each tweet. Finally, the top tweets in terms of cosine similarity with each claim from the reranked list were selected, resulting in a distinct set of 1,225 tweet-claim pairs with varying degrees of token and semantic similarity.

\begin{figure}[t]
    \centering
    \begin{tikzpicture}
        \node[draw, fill=mygray, rounded corners, drop shadow={fill=black!30, shadow xshift=3pt, shadow yshift=-3pt, opacity=0.5}, inner sep=10pt] {
            \begin{minipage}{.9\columnwidth}
                
                \begin{tabular}{l p{.9\columnwidth}}
                    \texttt{TWEET} & omg my dad got vaccinated yesterday and I just connected him to bluetooth \\
                    & \\
                    \texttt{CLAIM} & Vaccininated people emit Bluetooth signals. \\
                    & \\
                    \texttt{Question} & Which of the following options best describes the relationship between \texttt{TWEET} and \texttt{CLAIM}? \\
                    & \\
                    \texttt{Options} & If \texttt{TWEET} is true, then \texttt{CLAIM} is also true (entailment) \\
                    & If \texttt{TWEET} is true, then \texttt{CLAIM} cannot be said to be true or false (neutral) \\ 
                    & If \texttt{TWEET} is true, then \texttt{CLAIM} is false (contradiction).
                \end{tabular}
            
            \end{minipage}
        };
    \end{tikzpicture}
    \caption{Example of an entailment task instruction for human annotators}
    \label{fig:tweet_claim}
\end{figure}

\subsubsection{Human Annotation Task} \label{sec:annotation}

In line with previous work \cite{marelli2014}, human annotations for ground truth data were obtained through \textit{Amazon Mechanical Turk} (MTurk), an online platform for crowd-sourcing. To optimize the quality of our crowd-sourced data on MTurk, we specifically targeted top-tier workers. The filtering criteria included those identified as "MTurk Masters" by Amazon, with an approval rating exceeding 90\%, and located in the United States. For each task, we provided workers with instructions to classify each of the tweet-claim pairs into one of the options: If TWEET is true, then CLAIM is also true (\textit{entailment}); If TWEET is true, then CLAIM cannot be said to be true or false (\textit{neutral}); If TWEET is true, then CLAIM is false (\textit{contradiction}). An example of such task is shown in Figure \ref{fig:tweet_claim}. We also provided the annotators with three examples in the instructions, as illustrated in Figure \ref{fig:examples}.

\begin{figure}[h]
    \centering
    \begin{tikzpicture}
        \node[draw, fill=mygray, rounded corners, drop shadow={fill=black!30, shadow xshift=3pt, shadow yshift=-3pt, opacity=0.5}, inner sep=10pt] {
            \begin{minipage}{.9\columnwidth}
                \begin{tabular}{l p{.9\columnwidth}}
                    \texttt{TWEET} & A dog is running in a field. \\
                    \texttt{CLAIM} & An animal is running in a field. \\
                    \texttt{ANSWER} & A dog is an animal. A dog running in a field is an animal running in a field. So the final answer is ENTAILMENT. \\
                    & \\
                    \texttt{TWEET} & A man is breaking three eggs in a bowl. \\
                    \texttt{CLAIM} & A girl is pouring some milk in a bowl. \\
                    \texttt{ANSWER} & A man is breaking three eggs in a bowl does not imply that a girl is pouring some milk in a bowl. So the final answer is NEUTRAL. \\
                    & \\
                    \texttt{TWEET} & A man is playing golf. \\
                    \texttt{CLAIM} & No man is playing golf. \\
                    \texttt{ANSWER} & A man is playing golf and no man is playing golf cannot be true at the same time. So the final answer is CONTRADICTION.
                \end{tabular}
            \end{minipage}
        };
    \end{tikzpicture}
    \caption{Three examples used in entailment task instructions and few-shot-CoT prompts, derived from \cite{marelli2014}}
    \label{fig:examples}
\end{figure}

Because the presentation order matters in the entailment task, for tweet-claim pair, we acquired annotations from 5 different raters in the tweet-claim presentation order and also 5 in the claim-tweet order. For each pair of tweet-claim, the classifications for each of the presentation orders were determined by a majority vote scheme. We then labeled each tweet-claim pair as:

\begin{itemize}
  \item \textit{Entailment}, when the majority vote indicated so in either of the presentation orders.
  \item \textit{Contradiction}, when the majority vote indicated so in both presentation orders.
  \item \textit{Neutral}, when neither of the above two conditions was met.
\end{itemize}

When we evaluated models with this test set, we employed a more rigorous approach to account for possible biases and to produce a generalized assessment. Specifically, we generated 1,000 different combinations of tie-breakers and averaged the performance metrics across these combinations. Table \ref{tab:freq} provides a comprehensive summary of the class distributions within the test data, averaged across all generated combinations.

\begin{table}[ht]
\small
  \caption{Distribution of tweet-claim pairs for each entailment label.}
  \label{tab:freq}
  \begin{tabular}{lcc}
    \toprule
    Label & Count & Percentage  \\
    \midrule
    ENTAILMENT     & 791 & 64.6\% \\
    NEUTRAL        & 365 & 29.8\% \\
    CONTRADICTION  &  69 & 5.6\% \\
    \midrule
    TOTAL          & 1225 & 100\% \\
    \bottomrule
  \end{tabular}
\end{table}

\subsection{Pre-trained LLM Annotation}

To establish baselines, we compared the annotations across various pre-trained LLMs with human annotations. We used several LLMs, detailed in Table \ref{tab:model_description}, to assess their annotation capabilities. For consistency, only chat-based models were used. We set the temperature to 0 (or 0.01 for \textit{Llama} models) to ensure the annotation process was as deterministic as possible. Entailment task prompts, similar to the example shown in Figure \ref{fig:tweet_claim}, were fed to each LLM, and their responses were collected. Recognizing the importance of the presentation order in the entailment task, tweet-claim pairs were presented in both possible orders. After retrieving responses from the LLMs, we aggregated the results from both orders (cf., \S\ref{sec:annotation}).

LLMs' outputs are known to vary considerably depending on the prompts. Therefore, we tested the outputs from different prompting styles. We experimented with four distinct prompting styles. In the annotation-only setting, we prompted the LLMs to exclusively provide annotation results. In the zero-shot setting, LLMs were prompted to give explanations after providing the annotation results. In the zero-shot-CoT (\textit{chain-of-thought}) setting, multi-step reasoning was elicited from the LLMs by appending the request "\textit{Let's think step by step}" at the end of the prompt \cite{kojima2023} as shown in Figure \ref{fig:cot}. In the few-shot-CoT setting, LLMs were prompted to reason by providing three example pairs.

\begin{table}[t]
\small
\caption{Models utilized in this research.}
\label{tab:model_description}
\begin{tabular}{llll}
\toprule
Model & Avail. Date & Source & Producer \\
\midrule
\texttt{GPT-4} & Jul 6, 2023 & Proprietary & OpenAI \\
\texttt{GPT-3.5-Turbo} & Mar 1, 2023 & Proprietary & OpenAI \\
\texttt{Llama-2-70b} & Jul 18, 2023 & Partial Open & Meta \\
\texttt{Llama-2-13b} & Jul 18, 2023 & Partial Open & Meta \\
\texttt{Llama-2-7b} & Jul 18, 2023 & Partial Open & Meta \\
\bottomrule
\end{tabular}
\end{table}

\begin{figure}[h]
    \centering
    \begin{tikzpicture}
        \node[draw, fill=mygray, rounded corners, drop shadow={fill=black!30, shadow xshift=3pt, shadow yshift=-3pt, opacity=0.5}, inner sep=10pt] {
            \begin{minipage}{.9\columnwidth}
                \begin{tabular}{l p{.9\columnwidth}}
                    \texttt{System} & Which of the following best describes the relationship between TWEET and CLAIM? \\
                    & You must provide an explanation, and then a final choice as ENTAILMENT, NEUTRAL, or CONTRADICTION. \\
                    & \\
                    & If TWEET is true: \\
                    & (ENTAILMENT) then CLAIM is also true. \\
                    & (NEUTRAL) CLAIM cannot be said to be true or false. \\
                    & (CONTRADICTION) then CLAIM is false. \\
                    & \\
                    \texttt{Input} &  TWEET: Vaccininated people emit Bluetooth signals. \\
                    & CLAIM: omg my dad got vaccinated yesterday and I just connected him to bluetooth \\
                    & \\
                    \texttt{Output} & Firstly, the tweet is false because vaccinated people do not emit Bluetooth signals. This is a misinformation. The claim is based on the false information provided in the tweet. Therefore, if the tweet is true, then the claim could also be considered true. However, since the tweet is actually false, the claim is also false. Therefore, the relationship between the tweet and the claim is ENTAILMENT. \\
                \end{tabular}
            \end{minipage}
        };
    \end{tikzpicture}
    \caption{Example of an entailment task prompt in the zero-shot-CoT setting}
    \label{fig:cot}
\end{figure}

\begin{figure}[t]
    \centering
    \begin{tikzpicture}
        \node[draw, fill=mygray, rounded corners, drop shadow={fill=black!30, shadow xshift=3pt, shadow yshift=-3pt, opacity=0.5}, inner sep=10pt] {
            \begin{minipage}{.9\columnwidth}
                \begin{tabular}{l p{.9\columnwidth}}
                     & \textbf{Entailment} \\
                     & \\
                    \texttt{System} & Generate TWEET so that if TWEET is true, then CLAIM is also true. Be brief. Do not start a sentence with 'Just'. \\
                     & \\
                    \texttt{Input} & Vaccininated people emit Bluetooth signals. \\
                     & \\
                    \texttt{Output} & Crazy day. I'm fully vaccinated and now apparently I'm a walking Bluetooth signal! Get connected, folks! \#VaccineBluetooth \\
		      \end{tabular}
            \end{minipage}
            };
    \end{tikzpicture}

\vspace{.1cm}

    \begin{tikzpicture}
        \node[draw, fill=mygray, rounded corners, drop shadow={fill=black!30, shadow xshift=3pt, shadow yshift=-3pt, opacity=0.5}, inner sep=10pt] {
            \begin{minipage}{.9\columnwidth}
                \begin{tabular}{l p{.9\columnwidth}}
                     & \textbf{Contradiction} \\
                     & \\
                    \texttt{System} & Generate TWEET so that if TWEET is true, CLAIM is false. Be brief. Do not start a sentence with 'Just'. \\
                     & \\
                    \texttt{Input} & Vaccininated people emit Bluetooth signals. \\
                     & \\
                    \texttt{Output} & Had a full conversation with my vaccinated friend today, neither of our phones detected a Bluetooth signal. \#VaccineFacts \\
		      \end{tabular}
            \end{minipage}
            };
    \end{tikzpicture}

\vspace{.1cm}

    \begin{tikzpicture}
        \node[draw, fill=mygray, rounded corners, drop shadow={fill=black!30, shadow xshift=3pt, shadow yshift=-3pt, opacity=0.5}, inner sep=10pt] {
            \begin{minipage}{.9\columnwidth}
                \begin{tabular}{l p{.9\columnwidth}}
                     & \textbf{Neutral} \\
                     & \\
                    \texttt{System} & Generate TWEET so that if TWEET is true, then CLAIM is also true. Be brief. Do not start a sentence with 'Just'. \\
                     & \\
                    \texttt{Input} & Vaccininated people emit Bluetooth signals. \\
                     & \\
                    \texttt{Output} & Anybody encountered the strange phenomenon of finding a random Bluetooth signal on your device? So surreal! \#BluetoothMystery \\
		      \end{tabular}
            \end{minipage}
            };
    \end{tikzpicture}
    
    \caption{Examples of synthetic tweet generation prompts}
    \label{fig:gen_prompt}
\end{figure}

\subsection{Generating Synthetic Data and Fine-Tuning LLMs}
We generated synthetic training data with LLMs, as this process creates a balanced dataset tailored for specifics tasks, in this case claim matching. Fine-tuning language models optimizes them for task-specific nuances, possibly enhancing classification accuracy. Fine-tuning smaller models not only reduces computational costs for large-scale deployment but also allows for easier customization with newly emerging claims.

\subsubsection{Generating Train data sets}

To generate synthetic training data, we used three language models: \textit{GPT-4}, \textit{GPT-3.5-Turbo}, and \textit{Llama-2-70b-chat-hf}. Drawing from a collection of debunked claims, we engineered tweets that either supported, were neutral to, or contradicted these claims. We set the temperature parameter to 1 to facilitate the production of stylistically diverse outputs by language models. 
For a thorough evaluation, we exploited bidirectionality by generating two distinct types of synthetic tweets for each claim in both presentation orders. In the first presentation order, the models were prompted to generate tweets that either entail, contradict, or are neutral to a given claim. In the second presentation order, the models generated tweets that are either entailed, contradicted, or left neutral by the given claim. This approach ensured a comprehensive dataset that respects the importance of presentation order in textual entailment task. The specific prompts used for this data generation can be reviewed in Figure \ref{fig:gen_prompt}. 
In total, we generated 3,675 synthetic tweets for each of the three models and for each of the two presentation orders, resulting in a total of 22,050 tweets. This ensured a balanced distribution across the three categories of \textit{Entailment}, \textit{Contradiction}, and \textit{Neutral}.

\subsubsection{Fine-tuned LLM Annotation}

We fine-tuned \textit{GPT-3.5-Turbo}, \textit{Llama-2-13b-chat-hf}, and \textit{Llama-2-7b-chat-hf} using the training set described above. For our fine-tuning experiments, we divided the data into training and validation sets with an 80-20 split. The models were fine-tuned in both presentation orders: tweet-claim and claim-tweet, and we aggregated the output (cf., \S\ref{sec:annotation}). 
The fine-tuning approach varied by model. Specifically, \textit{GPT-3.5-Turbo} was fine-tuned via OpenAI's Fine-tuning API (\url{https://platform.openai.com/docs/guides/fine-tuning}), whereas \textit{Llama-2-13b-chat-hf} and \textit{Llama-2-7b-chat-hf} were fine-tuned using LoRA (Low-Rank Adaptation, \cite{hu2021}), through a Python framework \cite{llama-efficient-tuning}. Each model underwent three epochs of fine-tuning. Given the imbalanced nature of our task \cite{pado2022}, we also carried out experiments on an imbalanced training set, applying over and undersampling to achieve a distribution of 50\% for \textit{Entailment}, 35\% for \textit{Neutral}, and 15\% for \textit{Contradiction}. We performed fine-tuning and testing of the \textit{Llama} models on a single A100 GPU.

\section{Experiments}

To evaluate the efficacy of FACT-GPT, we performed two distinct sets of experiments. The first set examined the annotation results from various pre-trained models under four distinct prompting styles. The second set evaluated the performance of models fine-tuned on training sets generated from various LLMs. 
For the first experiment, we selected five pre-trained models: \textit{GPT-4}, \textit{GPT-3.5-Turbo}, \textit{Llama-2-70b-chat-hf}, \textit{Llama-2-13b-chat-hf}, and \textit{Llama-2-7b-chat-hf}. These models were tested in four prompting styles: annotation-only, zero-shot, zero-shot-CoT, and few-shot-CoT. To ensure more deterministic results, we set the temperature for each model at 0, or 0.01 for the \textit{Llama} models. This first experiment encompassed 20 distinct conditions. 
The second set of experiments involved fine-tuning three specific models: \textit{GPT-3.5-Turbo}, \textit{Llama-13b-chat-hf}, and \textit{Llama-7b-chat-hf}. We fine-tuned these models on training sets that were either balanced (1:1:1) or imbalanced (5:3.5:1.5) across three classes, generated from various pre-trained LLMs such as \textit{GPT-4}, \textit{GPT-3.5-Turbo}, and \textit{Llama-2-70b-chat-hf}. This second experiment consisted of 18 different conditions. The results from both sets of experiments reveal how various pre-trained and fine-tuned LLMs perform in claim matching tasks. 

\paragraph{Evaluation}

The models' outputs were compared with ground-truth annotations from human annotators. To quantify their effectiveness, we used various performance metrics such as (macro) precision, recall, and accuracy. These metrics revealed the strengths and weaknesses of the models in claim matching tasks. For the second set of experiments involving fine-tuning, we additionally monitored training loss at each step to track the models' learning progression. We also recorded validation loss and test performance at predetermined intervals, specifically every one-third of an epoch, to provide a fine-grained view of the models' performance over time. This allowed us to perform a detailed assessment of how quickly the models adapted to new data during the fine-tuning process, providing insights into their stability and robustness.

\subsection{Results}

\subsubsection{Pre-trained LLMs}

Table \ref{tab:pretrained} offers the results of the first experiment. While the assumption might be that \textit{GPT-4} would outperform other models in all metrics, our results indicate otherwise. While it did lead in annotation-only and few-shot recall, it did not universally outperform. In the annotation-only scenario, \textit{Llama-2-70b} actually had a higher precision and accuracy at .64 and .69, respectively. Moreover, \textit{GPT-3.5-Turbo} showed its strength in few-shot accuracy, scoring the highest at .67 while not sacrificing precision and recall too much when compared to \textit{GPT-4}. These results call into question the notion that a single model or approach can excel across all types of prompt styles in claim matching task. This variability in performance underscores the complexity of automated claim matching and serves as a caution against blindly selecting the largest models without a thorough evaluation. Ultimately, the data suggests that a more nuanced approach may be necessary for achieving optimal performance across diverse scenarios.

\subsubsection{Fine-tuned LLMs}

\begin{figure*}[h]
    \centering
    \includegraphics[width=1.0\textwidth]{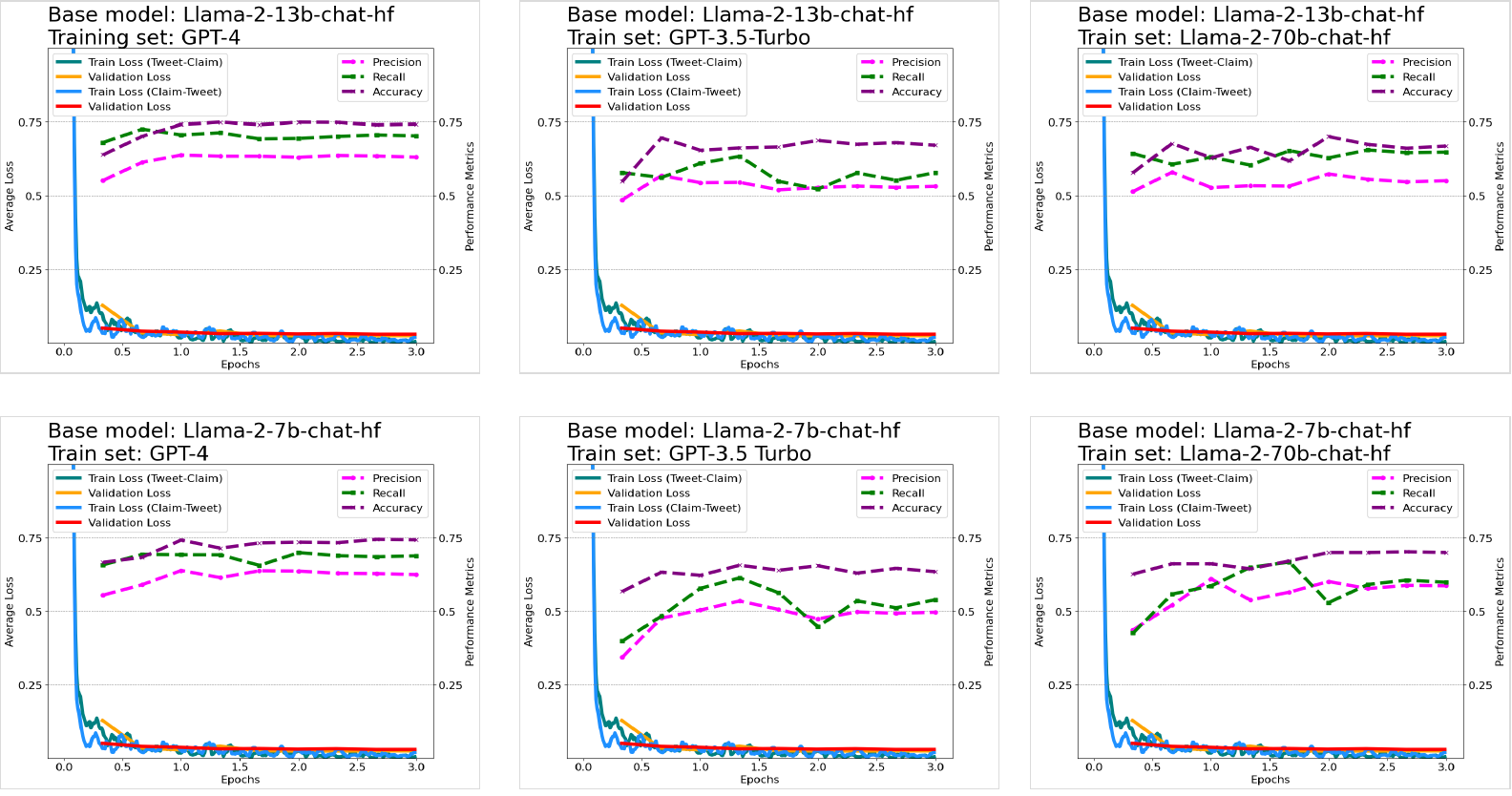}
    \caption{Train, validation loss and performance metrics for fine-tuned LLMs}
    \label{fig:loss_plot}
\end{figure*}

\begin{table}
\small
  \caption{Performance of pre-trained models.}
  \label{tab:pretrained}
  \begin{tabular}{llccc}
    \toprule
    Model & Prompt Style & Precison & Recall & Accuracy  \\
    \midrule
    \texttt{GPT-4}        & annotation-only & \textbf{.63} & .72 & \textbf{.65} \\
    \texttt{}                      & zero-shot       & .56 & .69 & .59 \\
    \texttt{}                      & zero-shot-CoT   & .58 & .72 & .59 \\
    \texttt{}                      & few-shot-CoT    & .61 & \textbf{.74} & .63 \\
    \midrule    
    \texttt{GPT-3.5-Turbo} & annotation-only & .55 & .57 & .58 \\
    \texttt{}                      & zero-shot       & .52 & .51 & .65 \\
    \texttt{}                      & zero-shot-CoT   & .54 & .65 & .63 \\
    \texttt{}                      & few-shot-CoT    & \textbf{.58} & \textbf{.71} & \textbf{.67} \\
    \midrule
    \texttt{Llama-2-70b}   & annotation-only & .64 & \textbf{.66} & \textbf{.69} \\
    \texttt{}                      & zero-shot       & .62 & .61 & .65 \\
    \texttt{}                      & zero-shot-CoT   & \textbf{.66} & .56 & .67 \\
    \texttt{}                      & few-shot-CoT    & .64 & .49 & .65 \\
    \midrule
    \texttt{Llama-2-13b}   & annotation-only & .48 & .54 & .35 \\
    \texttt{}                      & zero-shot       & \textbf{.50} & \textbf{.54} & .37 \\
    \texttt{}                      & zero-shot-CoT   & .48 & .49 & .61 \\
    \texttt{}                      & few-shot-CoT    & .47 & .41 & \textbf{.63} \\
    \midrule
    \texttt{Llama-2-7b}    & annotation-only & .25 & .33 & \textbf{.74} \\
    \texttt{}                      & zero-shot       & .40 & \textbf{.45} & .47 \\
    \texttt{}                      & zero-shot-CoT   & \textbf{.45} & .41 & .64 \\
    \texttt{}                      & few-shot-CoT    & .29 & .33 & .63 \\
    \bottomrule
    \\
    \\
    \\
  \end{tabular}
\end{table}

\begin{table}
\small
  \caption{Performance of fine-tuned models.}
  \label{tab:finetuned}
  \begin{tabular}{llccc}

    Balanced Set & (1:1:1) &&&\\
    \midrule
 Model               & Train Set           &   Precision &   Recall &   Accuracy \\
    \midrule
 \texttt{GPT-3.5-Turbo}       & \textit{GPT-4}                 &        \textbf{.66} &     \textbf{.67} &       \textbf{.78} \\
                     & \textit{GPT-3.5-Turbo}         &        .51 &     .57 &       .61 \\
                     & \textit{Llama-2-70b}           &        .58 &     .65 &       .69 \\
    \midrule
 \texttt{Llama-2-13b}         & \textit{GPT-4}                 &        \textbf{.63} &     \textbf{.70} &       \textbf{.74} \\
                     & \textit{GPT-3.5-Turbo}         &        .53 &     .58 &       .67 \\
                     & \textit{Llama-2-70b}           &        .55 &     .65 &       .67 \\
    \midrule
 \texttt{Llama-2-7b}          & \textit{GPT-4}                 &        \textbf{.62} &     \textbf{.69} &       \textbf{.74} \\
                     & \textit{GPT-3.5-Turbo}         &        .50 &     .54 &       .63 \\
                     & \textit{Llama-2-70b}           &        .59 &     .60 &       .70 \\
    \midrule
    &&&&\\
    Imbalanced Set & (5:3.5:1.5) &&&\\
    \midrule
 Model             & Train Set           &   Precision &   Recall &   Accuracy \\
    \midrule
 \texttt{GPT-3.5-Turbo}       & \textit{GPT-4}             &        \textbf{.65} &     \textbf{.70} &       \textbf{.77} \\
                  & \textit{GPT-3.5-Turbo}        &        .51 &     .54 &       .66 \\
                  & \textit{Llama-2-70b}          &        .55 &     .67 &       .66 \\
    \midrule
 \texttt{Llama-2-13b}      & \textit{GPT-4}                &        \textbf{.62} &     \textbf{.64} &       \textbf{.75} \\
                  & \textit{GPT-3.5-Turbo}        &        .53 &     .46 &       .70 \\
                  & \textit{Llama-2-70b}          &        .58 &     .61 &       .71  \\
    \midrule
 \texttt{Llama-2-7b}       & \textit{GPT-4}               &        \textbf{.61}  &     \textbf{.67} &       \textbf{.73} \\
                  & \textit{GPT-3.5-Turbo}       &        .51  &     .46 &       .68 \\
                  & \textit{Llama-2-70b}         &        .57 &     .56 &       .69 \\
    
    \bottomrule
\end{tabular}
\end{table}

In summary, the findings underscore the importance of the training set's quality and distribution for claim matching tasks, outweighing other factors such as model size or the class distribution of the training set. Moreover, when our models were fine-tuned using high-quality data generated by \textit{GPT-4}, they not only outperformed others but also reached peak performance more quickly and maintained this high level throughout the training process.

Table \ref{tab:finetuned} reveals significant findings from our second experiment. Specifically, smaller models fine-tuned on \textit{GPT-4}-generated sets exhibited comparable performance to their larger, pre-trained counterparts under ideal conditions. This outcome highlights the potential for more resource-efficient approaches in automated fact-checking.

When examining the performance of fine-tuned models, distinct patterns emerged. Three models—\textit{GPT-3.5-Turbo}, \textit{Llama-2-13b-chat-hf}, and \textit{Llama-2-7b-chat-hf}—excelled when fine-tuned on \textit{GPT-4}-generated training data. When trained on the same synthetic set, these models yielded similar results on a human-annotated test set. Moreover, these models exhibited only minor performance variations when trained on data sets with imbalanced classifications. These observations indicate that the quality of the training data plays a critical role in determining model performance.

Figure \ref{fig:loss_plot} further validates the robustness of these models fine-tuned on \textit{GPT-4}-generated training set. The data shows a consistent trend of stable training and validation loss across multiple epochs, confirming that the models are neither overfitting nor underfitting the data. Additionally, performance metrics such as accuracy, F1-score, and precision-recall curves also remained stable or showed gradual improvement over the epochs. This trend clearly stands out when compared with models trained on data synthesized with \textit{GPT-3.5-Turbo} or \textit{Llama-2-70b-chat-hf}, where performance metrics showed inconsistent fluctuation. Stability in metrics suggests that the models have effectively generalized to new, unseen data, corroborating their utility in real-world applications. The performance consistency across different evaluation measures affirms the resilience and reliability of these fine-tuned models, making them viable candidates for deployment in FACT-GPT pipeline.

\section{Discussion}
This work demonstrates the potential for large language models to augment the fact-checking workflow, particularly in the claim matching stage. Our results show that LLMs can reliably assess the relationships between social media posts and verified claims, offering performance comparable to human evaluations. This is consistent with the goals of augmented intelligence, which seeks to bolster human decision-making with informed AI recommendations \cite{murray2021}.

\paragraph{Limitations}
Our framework is naturally not immune from some limitations. Inference time for large, proprietary models may hinder real-time deployment, although smaller, domain-specific models could offer a more efficient alternative. The fact-checking process itself has inherent biases that are carried over into the training data for claim matching models. Fact checking is influenced by the priorities and choices of origin organizations, leading to collective blind spots around certain topics and political preferences \cite{rao2021political, jiang2021social}. The cross-referencing of topics across different media and fact-checking agencies is rare \cite{lim2018} due to logistical challenges and resource limitations. The fact-checking process can be influenced by the depth of scrutiny, the type of evidence used, and prior stances, often leaving decisions to individual media outlets \cite{rogerson2014}. Similarly to other machine learning systems, LLMs may propagate and even amplify societal and data-driven biases \cite{ferrara2023should, ferrara2023butterfly, ferrara2023genai}. Addressing these biases requires extensive human coordination.

Moving forward, maximizing AI benefits while mitigating risks requires ongoing collaboration among researchers, developers, and fact-checkers. All parties need to understand both the strengths and limitations of human and machine intelligence. A thoughtful implementation of claim matching and similar technologies can improve the fact-checkers’ ability to debunk misinformation, although human oversight and expertise remain indispensable.

\section{Related Work}

\subsection{Fact-checkers and Augmented Intelligence}

Fact-checkers play a crucial role in combating misinformation. Fact-checkers select public claims, gather multiple sources of evidence, and then verify or debunk these claims through logical analysis and expert consultations \cite{graves2019}. Over the years, they have established common practices and principles to ensure reliability \cite{ifcn2023}. These principles include non-partisanship, fairness, and transparency. As of 2022, the Duke Reporters' Lab identified 424 global fact-checking outlets, indicating a growth trend since 2014 \cite{stencel2023}. These outlets have scrutinized thousands of claims, creating vast datasets \cite{nakov2021automated}. Their true value, however, lies in consistently producing reliable information.

Integrating AI into the fact-checking process demands careful planning. The aim is to improve performance without disrupting established norms \cite{nakov2021automated}. While public sentiment towards AI is generally favorable in news coverage \cite{fast2017, chuan2019}, surveys \cite{nader2022}, and social media \cite{leiter2023}, concerns about its misuse for disseminating misinformation exist. Fact-checkers have expressed interest in AI tools for identifying claims and assessing their virality \cite{arnold2020}, but remain skeptical about AI completely replacing human judgment, emphasizing the irreplaceable aspect of human intuition.

The concept of 'augmented intelligence' appeals to fact-checkers. Rather than full automation, AI models that assist fact-checkers are more likely to gain acceptance. Services like Full Fact AI underscore AI's role as a helper, not a replacement. The broader AI community also advocates for empowering rather than replacing workers \cite{degallier2022}. Augmented intelligence aims to enhance human decision-making, not supplant it \cite{ieee2023}. While AI can offer predictive insights, it's crucial that these models also provide explanations for their recommendations, permitting human intervention when necessary \cite{murray2021}.

\subsection{Misinformation Detection}

Misinformation Detection (MID) is essential for studying the dissemination of false claims across diverse communication platforms. Researchers frequently use resources from fact-checkers to detect and analyze misinformation. The common method involves human annotation, employing keyword searches and manual tagging based on fact-checker guidelines. This approach is often favored for its accuracy but is labor-intensive and therefore not easily scalable.

Three primary methodologies are prevalent for MID in large-scale social media datasets:

\begin{itemize}
    \item \textbf{URL-based sampling}: Researchers rely on lists of untrustworthy websites, such as Zimdars’ 2016 document \cite{zimdars2016}, NewsGuard, and Media Bias/Fact Check, to identify questionable URLs \cite{sharma2022construction}. While efficient, this method has limitations, including missing tweets that lack URLs or failing to capture the linguistic features of false claims.
    \item \textbf{Hashtag-based sampling}: Particularly useful for politically sensitive topics, this method is efficient but risks capturing only a biased subset of posts \cite{rafail2018, chen2021}.
    \item \textbf{Keyword search}: Utilized by Ma et al. (2016), this method manually refines keywords extracted from fact-checked claims to yield relevant results \cite{ma2016}. It accounts for linguistic similarities but may involve arbitrary decision-making.
\end{itemize}

\begin{figure}[h]
    \centering
    \includegraphics[width=.4\textwidth]{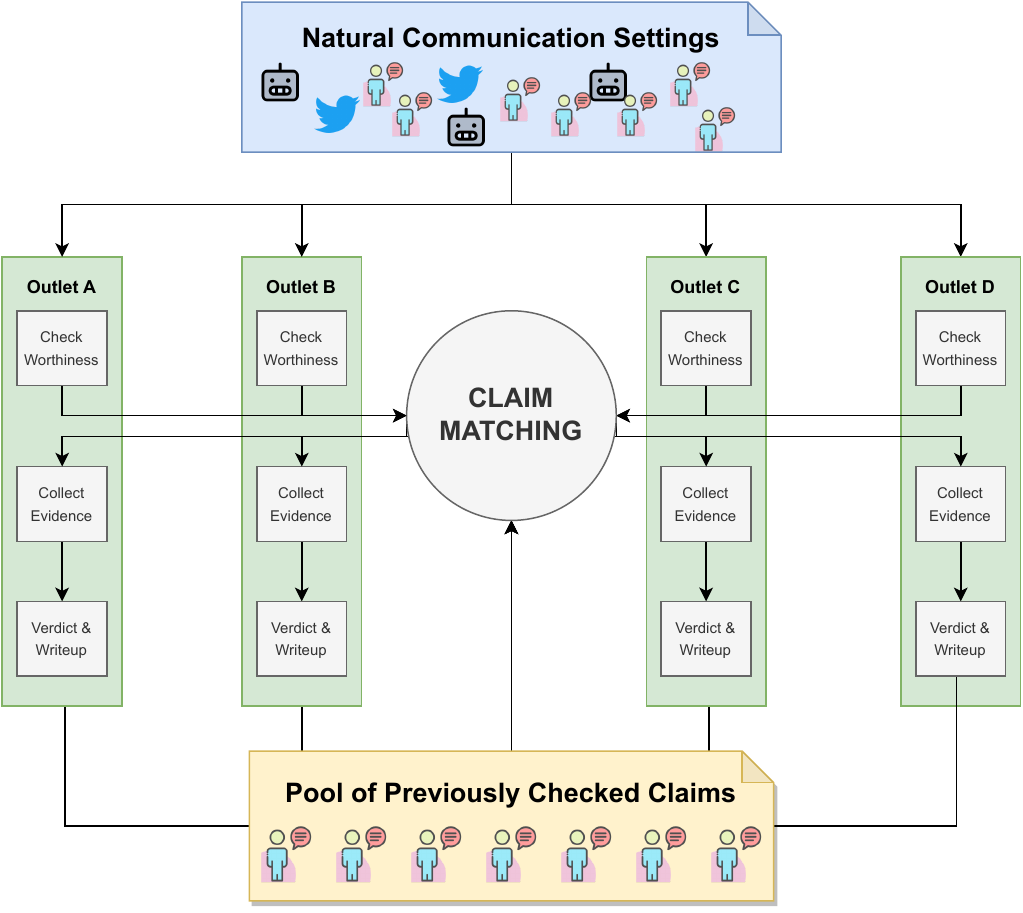}
    \caption{Collective process of claim matching}
    \label{fig:claim_matching}
\end{figure}

\textbf{Claim Matching} is a critical component in the Misinformation Detection (MID) workflow. It matches previously fact-checked claims with emerging claims from a variety of sources \cite{shaar2020}. The information verification pipeline, as conceptualized in prior research, outlines the various stages involved: assessing claim check-worthiness, claim matching, evidence retrieval, and claim factuality evaluation \cite{hassan2017, elsayed2019}. Claim matching models utilize both token and semantic similarities \cite{hassan2017, shaar2020}. As shown in Figure \ref{fig:claim_matching}, claim matching is a collective process that manages and leverages the pool of previously checked claims. The significance of claim matching arises from the propensity for false claims to be recycled and repeated in various forms \cite{nakov2021automated}. Efficient claim matching can facilitate early detection of misinformation, content moderation, and automated debunking \cite{he2023, vosoughi2018, ferrara2023social}.

\subsection{LLMs and Annotation Tasks}

LLMs have attracted considerable attention for their capability to automate a variety of annotation tasks. While platforms like Amazon Mechanical Turk (MTurk) facilitate crowd-sourced annotation, generating detailed datasets for complex tasks remains challenging \cite{dimaggio2015}. Due to their versatility, LLMs are under scrutiny to gauge how reliably they can handle the complexity of different annotation tasks. Studies have assessed LLMs in fact-checking \cite{hoes2023}, debunking cancer myths \cite{johnson2023}, annotating political tweets \cite{tornberg2023, gilardi2023}, and more. The generation of synthetic training data using GPT-based models to improve LLMs' classification task performance has also been investigated \cite{dai2023}.

Despite the promising avenues, it's crucial to recognize the inherent limitations of LLMs. Their proprietary nature makes understanding their decision-making challenging. Hoes et al. (2023) were unable to ascertain if the ChatGPT's fact-checking ability was inherent or due to data leakage \cite{hoes2023}. LLMs' probabilistic nature means their outputs can vary based on prompts and parameters \cite{reiss2023}. In comparative tests, ChatGPT often underperforms against finely-tuned, task-specific models \cite{kocon2023, zhu2023}. These results highlight LLMs' limitations in diverse settings.


\section{Conclusions}

This study demonstrates the potential for large language models (LLMs) to assist in the fact-checking workflow, specifically in the claim matching stage. Our findings suggest that LLMs can reliably judge the textual relationships between social media posts and verified claims. Properly fine-tuned smaller LLMs can perform comparably to much larger, proprietary models, offering more accessible and efficient AI solutions without sacrificing effectiveness.

Fully automating fact-checking with AI has risks and limitations. Biases can propagate through the models, and inconsistencies can arise from their probabilistic nature. Ongoing collaboration between researchers, developers, and practitioners is essential to maximize benefits while mitigating risks. With a well-planned and executed implementation strategy, claim matching technologies can be more effective in assisting fact-checkers by flagging false content at the initial stages. However, human oversight is vital as fact-checkers provide irreplaceable domain expertise.

Overall, this study shows the promise of claim matching models in offering fact-checkers informed recommendations about potentially misleading content. Our framework paves the way for future work integrating LLMs into the fact-checking pipeline. Using FACT-GPT to enhance fact-checkers aligns with the goals of augmented intelligence, which aims to empower human expertise through AI recommendations. Maintaining rigorous journalistic principles through human oversight is crucial to ensure the credibility and ethical integrity of the fact-checking process.

Moving forward, future studies should explore different strategies for data synthesis and data augmentation to improve FACT-GPT. Testing model reliability on diverse, real-world datasets is also needed. Research into the natural language explanation (NLE) of GPT models could enhance transparency \cite{huang2023}. This work offers a framework for using LLMs to assist human fact-checkers. Continued research and responsible AI development can empower fact-checkers to counter misinformation at scale.

\begin{acks}
\small This work was supported in part by DARPA (contract no. HR001121C0169).
\end{acks}

\balance
\bibliographystyle{ACM-Reference-Format}
\bibliography{sample-base}


\end{document}